\documentclass{article}
\usepackage{ijcai17}

\usepackage{times}

\usepackage{color}
\usepackage{graphicx}
\usepackage{ifthen}
\usepackage{array,multirow}
\usepackage{adjustbox}
\usepackage{comment}
\usepackage{algorithm}
\usepackage{algpseudocode}
\usepackage{amssymb}
\usepackage{amsmath}
\usepackage{subcaption}
\usepackage{hyperref}
\usepackage[font={small}]{caption}

\definecolor{gray}{rgb}{0.5,0.5,0.5} 
\definecolor{green}{rgb}{0, 0.4, 0} 
\definecolor{orange}{rgb}{1, 0.5, 0} 	
\definecolor{mahogany}{rgb}{0.75, 0.25, 0.0}
\definecolor{purple}{rgb}{0.6, 0, 0.6}
\definecolor{purple}{rgb}{0.6, 0, 0.6}
\definecolor{darkgreen}{rgb}{0, 0.4, 0.4} 
\definecolor{frenchblue}{rgb}{0.0, 0.45, 0.73}
\definecolor{what_color}{rgb}{0.7, 0.4, 0.3}

\newboolean{camera}
\setboolean{camera}{false}
\ifthenelse{\boolean{camera}}
{
	\newcommand{\yenchencamera}[2]		{\textcolor{orange}{#2}}
    \newcommand{\sunmincamera}[2]		{\textcolor{purple}{#2}}
        
} {
    \newcommand{\yenchencamera}[2]{#2}
	\newcommand{\sunmincamera}[2]{#2}
	
}

\newboolean{revising}
\setboolean{revising}{false}
\ifthenelse{\boolean{revising}}
{
	\newcommand{\ignore}[1]{}

	\newcommand{\yenchenreplace}[2]{\textcolor{darkgreen}{#2}}

	\newcommand{\sunmin}[1]{\textcolor{purple}{#1}}
	\newcommand{\sunminreplace}[2]{\textcolor{purple}{#2}}

} {
	\newcommand{\ignore}[1]{}

	\newcommand{\yenchenreplace}[2]{#2}

	\newcommand{\sunmin}[1]{#1}
	\newcommand{\sunminreplace}[2]{#2}

}

\title{Tactics of Adversarial Attack on Deep Reinforcement Learning Agents}
\author{Yen-Chen Lin$^1$, Zhang-Wei Hong$^1$, Yuan-Hong Liao$^1$, Meng-Li Shih$^1$, Ming-Yu Liu$^2$, Min Sun$^1$\\
$^1$National Tsing Hua University, Taiwan
\\$^2$NVIDIA, Santa Clara, California, USA\\
\{yenchenlin@gapp, williamd4112@gapp, s102061137@m102, shihsml@gapp, sunmin@ee\}.nthu.edu.tw
\\ mingyul@nvidia.com
}

\begin{document}

\maketitle

\begin{abstract}
We introduce two tactics, namely the strategically-timed attack and the enchanting attack, to attack reinforcement learning agents trained by deep reinforcement learning algorithms using adversarial examples. In the strategically-timed attack, the adversary aims at minimizing the agent's reward by only attacking the agent at a small subset of time steps in an episode. Limiting the attack activity to this subset helps prevent detection of the attack by the agent. We propose a novel method to determine when an adversarial example should be crafted and applied. In the enchanting attack, the adversary aims at luring the agent to a designated target state. This is achieved by combining a generative model and a planning algorithm: while the generative model predicts the future states, the planning algorithm generates a preferred sequence of actions for luring the agent. A sequence of adversarial examples is then crafted to lure the agent to take the preferred sequence of actions. We apply the proposed tactics to the agents trained by the state-of-the-art deep reinforcement learning algorithm including DQN and A3C. In 5 Atari games, our strategically-timed attack reduces as much reward as the uniform attack \sunmin{(i.e., attacking at every time step)} does by attacking the agent 4 times less often. Our enchanting attack lures the agent toward designated target states with a more than 70\% success rate. \sunmincamera{}{Example videos} are available at \url{http://yenchenlin.me/adversarial_attack_RL/}.
\end{abstract}

\section{Introduction}\label{sec.Intro}

Deep neural networks (DNNs), which can extract hierarchical distributed representations from signals, are established as the de facto tool for pattern recognition, particularly for supervised learning. We, as a generation, have witnessed a trend of fast adoption of DNNs in various commercial systems performing image recognition~\cite{krizhevsky2012imagenet}, speech recognition~\cite{hannun2014deep}, and natural language processing~\cite{sutskever2014sequence} tasks. Recently, DNNs have also started to play a central role in reinforcement learning (RL)---a field of machine learning research where the goal is to train an agent to interact with the environment for maximizing its reward. The community has realized that DNNs are ideal function approximators for classical RL algorithms, because DNNs can extract reliable patterns from signals for constructing a more informed action determination process. For example, \cite{mnih:human} use a DNN to model the action--value function in the Q-learning algorithm, and \cite{mnih:asynchronous} use a DNN to directly model the policy. Reinforcement learning research powered by DNNs is generally referred to as deep reinforcement learning (Deep RL).

However, a constant question lingers while we enjoy using DNNs for function approximation in RL. Specifically, since DNNs are known to be vulnerable to the adversarial example attack~\cite{szegedy:intriguing}, as a deep RL agent inherits the pattern recognition power from a DNN, does it also inherit its vulnerability to the adversarial examples? We believe the answer is yes and provide empirical evidence in the paper. 

Adversarial attack on deep RL agents is different from adversarial attack on classification system in several ways. Firstly, \yenchencamera{a deep RL agent makes sequential decisions, an adversarial attack at a time instance affects all subsequent decisions.}{an RL agent interacts with the environment through a sequence of actions where each action changes the state of the environment. What the agent received is a sequence of correlated observations. For an episode of $L$ steps, an adversary can determine whether to craft an adversarial example to attack the agent at each time step (i.e. there are $2^L$ choices).} 
Secondly, an adversary to deep RL agents have different goals such as reducing the final rewards of agents or malevolently lure agents to dangerous states, which is different to an adversary to classification system that aims at lowering classification accuracy. In this paper, we focus on studying adversarial attack specific on deep RL agents. We argue this is important. As considering deep RL agents for controlling machines, we need to understand the vulnerability of the agents because it would limit their use in mission-critical tasks such as autonomous driving. Based on~\cite{kurakin:adversarial}, which showed that adversarial examples also exist in the real world, an adversary can add maliciously-placed paint to the surface of a traffic stop to confuse an autonomous car. How could we fully trust deep RL agents if their vulnerability to adversarial attacks is not fully understood and addressed?

\yenchencamera{The adversarial attack on a deep RL agent is different to that on a DNN-based classification system.In RL, an agent interacts with the environment through a sequence of actions where each action changes the state of the environment. What the agent received is a sequence of correlated observations. For an episode of $L$ steps, an adversary can determine whether to craft an adversarial example to attack the agent at each time step (i.e. there are $2^L$ choices).}{} In a contemporary work,~\cite{sandy:adversarial} proposes an adversarial attack tactic where the adversary attacks a deep RL agent at every time step in an episode. We refer to such a tactic as the uniform attack and argue it is preliminary. First, the uniform attack ignores the fact that the observations are correlated. Moreover, the spirit of adversarial attack is to apply a minimal perturbation to the observation to avoid detection. If the adversary perturbs the observation at every time instance, it is more likely to be detected. A more sophisticated strategy would be to attack at selective time steps. For example, as shown in Fig.~\ref{fig::timed_attack}, attacking the deep RL agent has no consequence when the ball is far away from the paddle. However, when the ball is close to the paddle, attacking the deep RL agent could cause it to drop the ball. Therefore, the adversarial attacks at different time instances are not equally effective. Based on this observation, we propose the strategically-timed attack, which takes into account the number of times an adversarial example is crafted and used. It intends to reduce the reward with as fewer adversarial examples as possible. An adversarial example is only used when the attack is expected to be effective. Our experiment results show that an adversary exercising the strategically-timed attack tactic can reduce the reward of the state-of-the-art deep RL agents by attacking four times less often as comparing to an adversary exercising the uniform attack tactic.

In addition, we propose the enchanting attack for maliciously luring a deep RL agent to a certain state. While the strategically-timed attack aims at reducing the reward of a deep RL agent, the enchanting attack aims at misguiding the agent to a specified state. The enchanting attack can be used to mislead a self-driving car controlled by a deep RL agent to hit a certain obstacle. We implement the enchanting attack using a planning algorithm and a deep generative model. To the best of our knowledge, this is the first planning-based adversarial attack on a deep RL agent. Our experiment results show that the enchanting attack has a more than $70\%$ success rate in attacking state-of-the-art deep RL agents.

We apply our adversarial attack to the agents trained by state-of-the-art deep RL algorithms including A3C~\cite{mnih:asynchronous} and DQN~\cite{mnih:human} on 5 Atari games. We provide examples to evaluate the effectiveness of our attacks. We also compare the robustness of the agents trained by the A3C and DQN algorithms to these adversarial attacks. The contributions of the paper are summarized below:
\begin{itemize}
\item We study adversarial example attacks on deep RL agents trained by state-of-the-art deep RL algorithms including A3C and DQN.
\ignore{\item We propose the strategically-timed attack and the enchanting attack. While the strategically-timed attack aims at attacking a deep RL agent at critical moments, the enchanting attack aims at maliciously luring an agent to a certain state.}
\item \sunmin{We propose the strategically-timed attack aiming at attacking a deep RL agent at critical moments.}
\item \sunmin{We propose the enchanting attack (the first planning-based adversarial attack) aiming at maliciously luring an agent to a certain state.}
\item We conduct extensive experiments to evaluate the vulnerability of deep RL agents to the two attacks.
\end{itemize}

\section{Related Work}\label{sec.Background}

Following~\cite{szegedy:intriguing}, several adversarial example generation methods were proposed for attacking DNNs. Most of these methods generated an adversarial example via seeking a minimal perturbation of an image that can confuse the classifier (e.g., ~\cite{goodfellow:explaining,kurakin:adversarial}). \cite{moosavi:deep} first estimated linear decision boundaries between classes of a DNN in the image space and iteratively shifted an image toward the closest of these boundaries for crafting an adversarial example.

While the existence of adversarial examples to DNNs has been demonstrated several times on various supervised learning tasks, the existence of adversarial examples to deep RL agents has remained largely unexplored. In a contemporary paper,~\cite{sandy:adversarial} proposed the uniform attack, which attacks a deep RL agent with adversarial examples at every time step in an episode for reducing the reward of the agent. Our work is different to~\cite{sandy:adversarial} in several aspects, including 1) we introduce a strategically-timed attack, which can reach the same effect of the uniform attack by attacking the agent four times less often on average; 2) we also introduce an enchanting attack tactic, which is the first planning-based adversarial attack to misguide the agent toward a target state.

In terms of defending DNNs from adversarial attacks, several approaches were recently proposed. \cite{goodfellow:explaining} augmented the training data with adversarial examples to improve DNNs' robustness to adversarial examples. \cite{zheng:improving} proposed incorporating a stability term to the objective function, encouraging DNNs to generate similar outputs for various perturbed versions of an image. Defensive distillation is proposed in~\cite{papernot:distillation} for training a network to defend both the L-BFGS attack in~\cite{szegedy:intriguing} and the fast gradient sign attack in~\cite{goodfellow:explaining}. Interestingly, as anti-adversarial attack approaches were proposed, stronger adversarial attack approaches also emerged. \cite{carlini-wagner:towards} recently introduced a way to construct adversarial examples that is immune to various anti-adversarial attack methods, including defensive distillation. A study in~\cite{rozsa:towards} showed that more accurate models tend to be more robust to adversarial examples, while adversarial examples that can fool a more accurate model can also fool a less accurate model. As the study of adversarial attack to deep RL agents is still in its infancy, we are unaware of earlier works on the anti-adversarial attack to deep RL agents.

\section{Adversarial Attacks}\label{sec.Attack}

\begin{figure*}[t!]
\centering
\includegraphics[trim=0.0in 0.0in 0.0in 0in,width=0.99\textwidth]{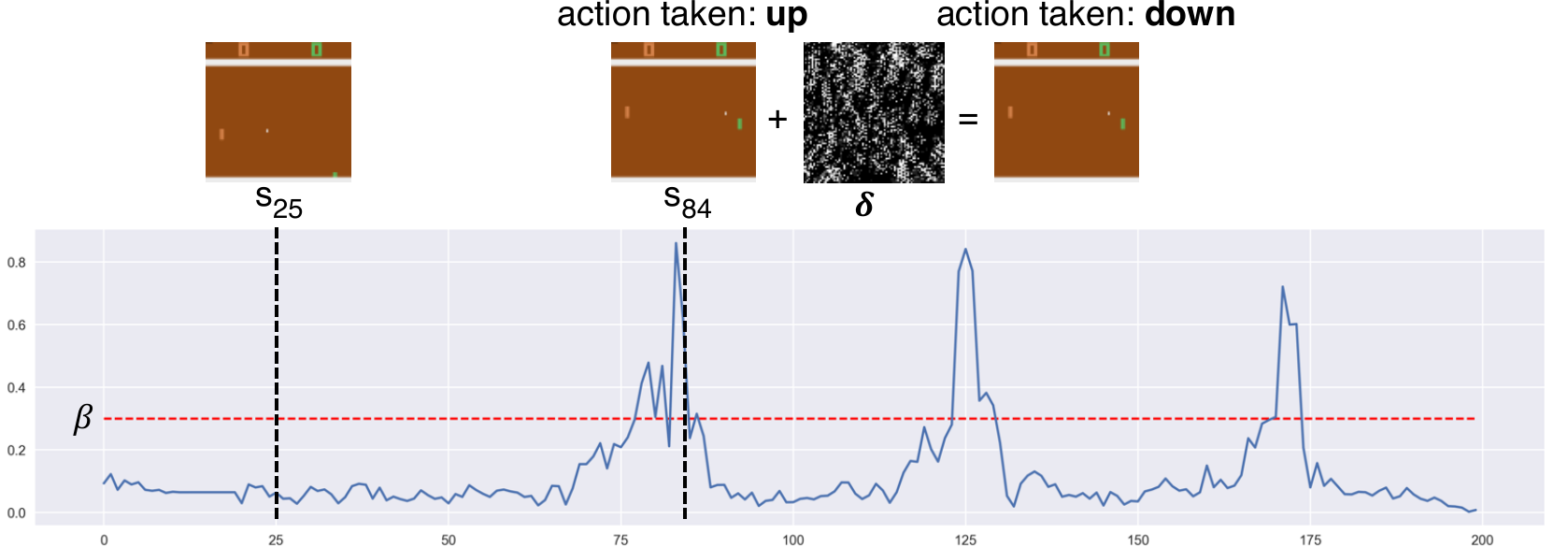}
\caption{Illustration of the strategically-timed attack on Pong. We use a function $c$ to compute the preference of the agent in taking the most preferred action over the least preferred action at the current state $s_t$. A large preference value implies an immediate reward. In the bottom panel, we plot $c(s_t)$. Our proposed strategically-timed attack launch an attack to a deep RL agent when the preference is greater than or equal to a threshold, $c(s_t)\ge\beta$ (red-dash line). When a small perturbation is added to the observation at $s_{84}$ (where $c(s_{84})\ge\beta$), the agent changes its action from up to down and eventually misses the ball. But when the perturbation is added to the observation at $s_{25}$ (where $c(s_{25})<\beta$), there is no impact to the reward.}\label{fig::timed_attack}
\end{figure*}

In this section, we will first review the adversarial example attack to DNN-based classification systems. We will then generalize the attack to deep RL agents and introduce our strategically-timed and enchanting attacks.

\subsection{Preliminaries}

Let $x$ be an image and $f$ be a DNN. An adversarial example to the DNN can be crafted through solving the following optimization problem:
\begin{align}
&\min_{\delta} \mathcal{D}_I(x,x+\delta)\nonumber\\
&\quad\text{subject to}\quad f(x)\neq f(x+\delta),
\label{eqn::adversarial_attack_an_image}
\end{align}
where $\mathcal{D}_I$ is an image similarity metric. In words, it looks for a minimal perturbation, $\delta$, of an image that can change the class assignment of the DNN to the image.

An RL agent learns to interact with the environment through the rewards signal. At each time step, it performs an action based on the observation of the environment for maximizing the accumulated future rewards. The action determination is through a policy $\pi$, which maps an observation to an action. Let the current time step be $t$, the goal of an RL algorithm is then to learn a policy that maximizes the accumulated future rewards $\mathcal{R}_t = r_t + r_{t+1}+...+r_{L}$, where $L$ is the length of an episode. 

In a deep RL algorithm, the policy $\pi$ is modeled through a DNN. An adversary can attack an agent trained by the deep RL algorithm by perturbing the observations (through crafting an adversarial example) to make the agent take non-preferred actions that can result in reduction of the accumulated future rewards.


\subsection{Adversarial Attacks on RL}

In a recent paper,~\cite{sandy:adversarial} propose the uniform attack tactic where the adversary attacks a deep RL agent at every time step, by perturbing each image the agent observes. The perturbation to an image is computed by using the fast gradient sign method~\cite{goodfellow:explaining}. The uniform attack tactic is regarded as a direct extension of the adversarial attack on a DNN-based classification system, since the adversarial example at each time step is computed independently of the adversarial examples at other time steps. 

It does not consider several unique aspects of the RL problem. For example, during learning, an RL agent is never told which actions to take but instead discovers which actions yield the most reward. This is in contrast to the classification problem where each image has a ground truth class. Moreover, an adversarial attack to a DNN is considered a success if it makes the DNN outputs any wrong class. But the success of an adversarial attack on an RL agent is measured based on the amount of reward that the adversary takes away from the RL agent. Instead of perturbing the image to make the agent takes any non-optimal action, we would like to find a perturbation that makes the agent takes an action that can reduce most reward. Also, because the reward signal in many RL problems is sparse, an adversary need not attack the RL agent at every time step. Our strategically-timed attack tactic described in Section~\ref{adversarial_rl} leverages these unique characteristics to attack deep RL agents.

Another unique characteristic of the RL problem is that each action taken by the agent influenced its future observations. Therefore, an adversary could plan a sequence of adversarial examples to maliciously lure the agent toward a certain state that can lead to a catastrophic outcome. Our enchanting attack tactic described in Section~\ref{sec.State_Attack} leverages this characteristic to attack RL agents.

\subsection{Strategically-Timed Attack}\label{adversarial_rl}

In an episode, an RL agent observes a sequence of observations or states $\{s_{1},...,s_{L}\}$. Instead of attacking at every time step in an episode, the strategically-timed attack selects a subset of time steps to attack the agent. Let $\{\delta_{1},...,\delta_{L}\}$ be a sequence of perturbations. Let $\mathcal{R}_1$ be the expected return at the first time step. We can formulate the above intuition as an optimization problem as follows:
\begin{align}
\min_{b_1,b_2,...,b_L,\delta_1,\delta_2,...,\delta_L} & R_{1}(\bar{s}_{1},...,\bar{s}_{L}) &\nonumber\\
&\bar{s}_t = s_t + b_t \delta_t \quad &\text{for all } t=1,...,L\nonumber\\
&b_t \in \{0, 1\}, \quad &\text{for all } t=1,...,L\nonumber\\
&\sum_t b_t \leq \Gamma &
\label{eq.p-attack}
\end{align}
The binary variables $b_1,...,b_L$ denote when an adversarial example is applied. If $b_t=1$, the perturbation $\delta_t$ is applied. Otherwise, we do not alter the state. The total number of attacks is limited by the constant $\Gamma$. In words, the adversary minimizes the expected accumulated reward by strategically attacking less than $\Gamma << L$ time steps.

The optimization problem in~(\ref{eq.p-attack}) is a mixed integer programming problem, which is difficult to \yenchencamera{solved}{solve}. Moreover, in an RL problem, the observation at time step $t$ depends on all the previous observations, which makes the development of a solver to~(\ref{eq.p-attack}) even more challenging \yenchencamera{}{since the problem size grows exponentially with $L$}. In order to study adversarial attack to deep RL agents, we bypass these limitations and propose a heuristic algorithm to compute $\{b_1,...,b_L\}$ (solving the when-to-attack problem) and $\{\delta_{1},...,\delta_{L}\}$ (solving the how-to-attack problem), respectively. In the following, we first discuss our solution to the when-to-attack problem. We then discuss our solution to the how-to-attack problem.

\subsubsection{When to attack}\label{when}

We introduce a relative action preference function $c$ for solving the when-to-attack problem. The function $c$ computes the preference of the agent in taking the most preferred action over the least preferred action at the current state $s_t$ \yenchencamera{}{(similar to~\cite{amir:action-gap}).} The degree of preference to an action depends on the DNN policy. A large $c$ value implies that the agent strongly prefers one action over the other. In the case of Pong, when the ball is about to drop from the top of the screen (see $s_{84}$ in Fig.~\ref {fig::timed_attack}), a well-trained RL agent would strongly prefer an up action over a down action. But when the ball is far away from the paddle (see $s_{25}$ in Fig.~\ref {fig::timed_attack}), the agent has no preference on any actions, resulting a small $c$ value. We describe how to design the relative action preference function $c$ for attacking the agents trained by the A3C and DQN algorithms below.

For policy gradient-based methods such as the A3C algorithm, if the action distribution of a well-trained agent is uniform at state $s_t$, it means that taking any action is equally good. But, when an agent strongly prefer a specific action (The action has a relative high probability.), it means that it is critical to perform the action; otherwise the accumulated reward will be reduced. Based on this intuition, we define the $c$ function as 
\begin{align}
c(s_t) = \max_{a_t} \pi (s_t, a_t) - \min_{a_t} \pi (s_t, a_t).
\end{align}
where $\pi$ is the policy network which maps a state--action pair $(s_t,a_t)$ to a probability, representing the likelihood that the action $a_t$ is chosen. In our strategically-timed attack, the adversary attacks the deep RL agent at time step $t$ when the relative action preference function $c$ has a value greater a threshold parameter $\beta$. In other words, $b_t=1$ if and only if $c(s_t)\ge\beta$. We note the $\beta$ parameter controls how often it attacks the RL agent and is related to $\Gamma$.

For value-based methods such as DQN, the same intuition applies. We can convert the computed Q-values of actions into a probability distribution over actions using the softmax function (similar to~\cite{sandy:adversarial}) with the temperature constant $T$. 
\begin{align}
c(s_t) = \max_{a_t} \frac{e^{\frac{Q(s_t,a_t)}{T}}}{\sum_{a_k}e^{\frac{Q(s_t,a_k)}{T}}} - \min_{a_t} \frac{e^{\frac{Q(s_t,a_t)}{T}}}{\sum_{a_k}e^{\frac{Q(s_t,a_k)}{T}}}.
\label{eq.convert}
\end{align}

\subsubsection{How to attack}\label{how}

To craft an adversarial example at time step $\delta_t$, we search for a perturbation to be added to the observation that can change the preferred action of the agent from the originally (before applying the perturbation) most preferred one to the originally least preferred one. We use the attack method introduced in \cite{carlini-wagner:towards} where we treat the least-preferred action as the misclassification target (see Sec.~\ref{sec.exp_setup} for details). This approach allows us to leverage the output of a trained deep RL agent as cue to craft effective adversarial example for reducing accumulated rewards.
\subsection{Enchanting Attack}\label{sec.State_Attack}

\begin{figure*}[t!]\vspace{-2mm}
\begin{center}
\includegraphics[width=1.000\linewidth]{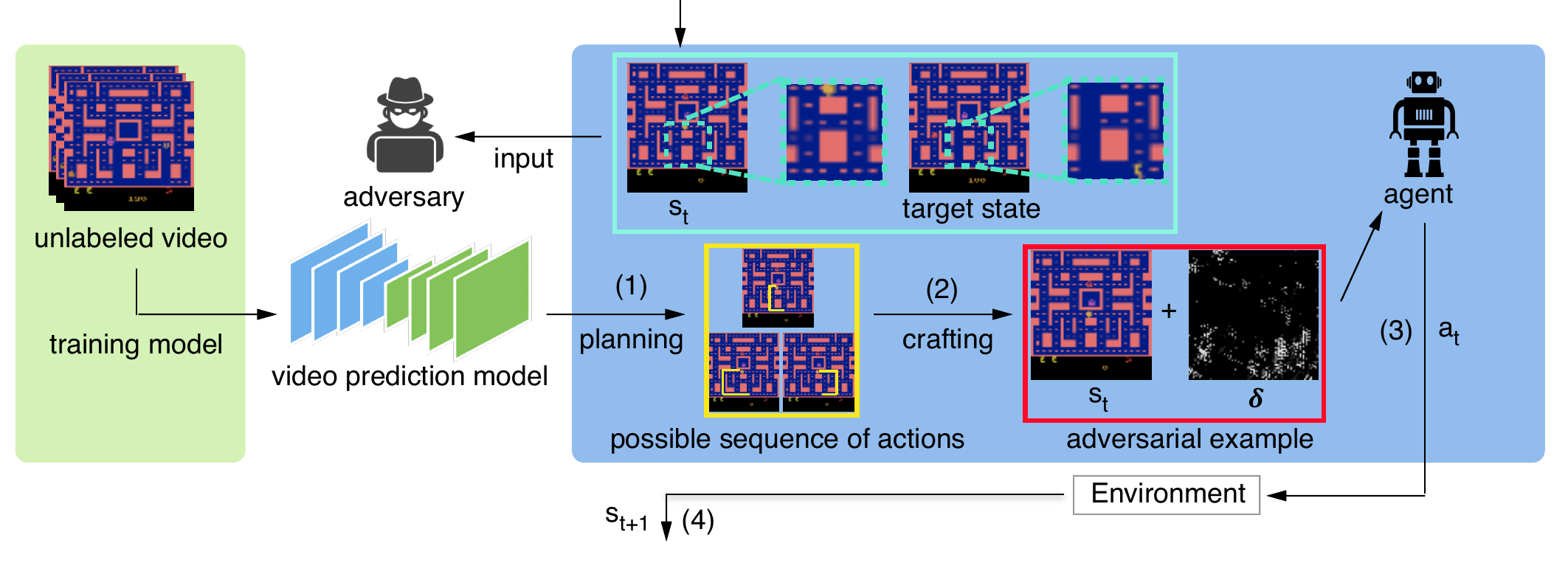}
\end{center}
\vspace{-7mm}
\caption{Illustration of Enchanting Attack on Ms.Pacman. The blue panel on the right shows the flow of the attack starting at $s_t$: (1) action sequence planning, (2) crafting an adversarial example with a target-action, (3) the agent takes an action, and (4) environment generates the next state $s_{t+1}$. The green panel at the left depicts that the video prediction model is trained from unlabeled video. The white panel in the middle depicts the adversary starts at $s_t$ and utilize the prediction model to plan the attack.}
\label{fig.state_attack_flow}
\end{figure*}

The goal of the enchanting attack is to lure the deep RL agent from current state $s_{t}$ at time step $t$ to a specified target state $s_{g}$ after $H$ steps. The adversary needs to craft a series of adversarial examples $s_{t+1}+\delta_{t+1},...,s_{t+H}+\delta_{t+H}$ for this attack. The enchanting attack is therefore more challenging than the strategically-timed attack.

We break this challenging task into two subtasks. In the first subtask, we assume that we have full control of the agent to take arbitrary actions at each step. Hence, the task is reduced to planning a sequence of actions for reaching the target state $s_g$ from current state $s_t$. In the second subtask, we craft an adversarial example $s_t+\delta_t$ to lure an agent to take the first action of planned action sequence using the method introduced in~\cite{carlini-wagner:towards}. After the agent observes the adversarial example and takes the first action planned by the adversary, the environment will return a new state $s_{t+1}$. We progressively craft $s_{t+1}+\delta_{t+1},...,s_{t+H}+\delta_{t+H}$, one at a time, using the same procedure described in (Fig.~\ref{fig.state_attack_flow}) to lure the agent from state $s_{t+1}$ to the target state $s_g$. Next, we describe an on-line planing algorithm, which makes use of a next frame prediction model, for generating the planned action sequence.

\subsubsection{Future state prediction and evaluation}

We train a video prediction model $M$ to predict a future state given a sequence of actions based on~\cite{oh:action}, which used a generative model to predict a video frame in the future:
\begin{eqnarray}
s_{t+H}^{M} = M(s_{t}, A_{t:t+H})~,\label{eq.M}
\end{eqnarray}
where $A_{t:t+H}=\{a_{t},...,a_{t+H}\}$ is the given sequence of $H$ future actions beginning at step $t$, $s_t$ is the current state, and $s_{t+H}^{M}$ is the predicted future state. For more details about the video prediction model, please refer to the original paper.

The series of actions $A_{t:t+H}=\{a_{t},...,a_{t+H}\}$ take the agent to reach the state $s_{t+H}^{M}$. Since the goal of the enchanting attack is to reach the target state $s_g$, we can evaluate the success of the attack based on the \sunminreplace{similarity}{distance} between $s_g$ and $s_{t+H}^{M}$, which is given by $D(s_{g}, s_{t+H}^{M})$. The \sunminreplace{similarity}{distance}  $D$ is realized using the $L_2$-norm in our experiments. We note that other metrics can be applied as well. We also note that the state is given by the observed image by the agent.

\subsubsection{Sampling-based action planning}

\yenchenreplace{
We use the cross-entropy method (CEM) ~\cite{rubinstein:cross} to compute a sequence of actions to steer the RL agent toward our target state. Specifically, we first create a distribution of state-transition models where the distribution is modeled by a two layer neural network. At each iteration, we sample $N$ state-transition models from the distribution. Each model produces an action sequence of length $H$: $\{A^{n}_{t:t+H}\}_{n=1}^{N}$. We rank the model based on the distance between the final state obtained by applying the model and the target state $s_g$. After that, the $K$-best models are identified, and the distribution is updated based on the parameters of the $K$-best trajectories. We repeat the above process for $J$ times. In our experiments, the hyper-parameter values are $N=200$, $K=40$, and $J=5$.}{}
We use a sampling-based cross-entropy method (~\cite{rubinstein:cross}) to compute a sequence of actions to steer the RL agent toward our target state. Specifically, we sample $N$ action sequences of length $H$: $\{A^{n}_{t:t+H}\}_{n=1}^{N}$, and rank each of them based on the distance between the final state obtained after performing the action sequence and the target state $s_g$. After that, we keep the best $K$ action sequences and refit our categorical distributions to them. In our experiments, the hyper-parameter values are $N=2000$, $K=400$, and $J=5$.

At the end of the last iteration, we set the sampled action sequence $A^{*}_{t:t+H}$ that results in a final state that is \sunminreplace{most close}{closest} to our target state $s_g$ as our plan. Then, we craft an adversarial example with the target-action $a^*_t$ using the method introduced in~\cite{carlini-wagner:towards}. Instead of directly crafting the next adversarial example with target-action $a^*_{t+1}$, we plan for another enchanting attack starting at state $s_{t+1}$ to be robust to \sunmin{potential} failure in the previous attack.

We note that the state-transition model is different to the policy of the deep RL agent. We use the state-transition model to propose a sequence of actions that we want the deep RL agent to follow. We also note that both the state-transition model and the future frame prediction model $M$ are learned without assuming any information from the RL agent.

\section{Experiments}\label{sec.Exp}

We evaluated our tactics of adversarial attack to deep RL agents on 5 different Atari 2600 games (i.e., MsPacman, Pong, Seaquest, Qbert, and ChopperCommand) using OpenAI Gym~\cite{brockman:gym}. These games represents a balanced collection. Deep RL agents can achieve an above-human level performance when playing Pong and a below-human level performance when playing Ms.Pacman. We discuss our experimental setup and results in details. Our implementation will be released.

\subsection{Experimental Setup}\label{sec.exp_setup}

For each game, the deep RL agents were trained using the state-of-the-art deep RL algorithms including the A3C and DQN algorithms. For the agents trained by the A3C algorithm, we used the same pre-processing steps and neural network architecture as in~\cite{mnih:asynchronous}. For the agents trained by the DQN algorithm, we also used the same network architecture for the Q-function as in the original paper~\cite{mnih:human}. The input to the neural network at time $t$ was the concatenation of the last 4 images. Each of the images was resized to $84 \times 84$. The pixel value was rescaled to \yenchenreplace{$[0\quad 1]$}{$[0, 1]$}. The output of the policy was a distribution over possible actions for A3C, and an estimate of Q values for DQN.

Although several existing methods can be used to craft an adversarial example (e.g., the fast gradient sign method~\cite{goodfellow:explaining}, and Jacobian-based saliency map attack~\cite{papernot:limitations}), anti-adversarial attack measures were also discovered to limit their impact ~\cite{goodfellow:explaining,papernot:distillation}. We adopted an adversarial example crafting method proposed by~\cite{carlini-wagner:towards}, which can break several existing anti-adversarial attack methods. Specifically, it crafts an adversarial example by approximately optimizing~(\ref{eqn::adversarial_attack_an_image}) where the image similarity metric was given by $L_{\infty}$ norm. We early stopped the optimizer when $\mathcal{D}(s, s + \delta)< \epsilon$, where $\epsilon$ is a small value set to $0.007$. \yenchencamera{}{The value of temperature $T$ in Equation (\ref{eq.convert}) is set to $1$ in the experiments.}

\begin{figure*}[h!]\vspace{-2mm}
\centering
\begin{subfigure}{.2\textwidth}
  \centering
  \includegraphics[width=1.0\linewidth]{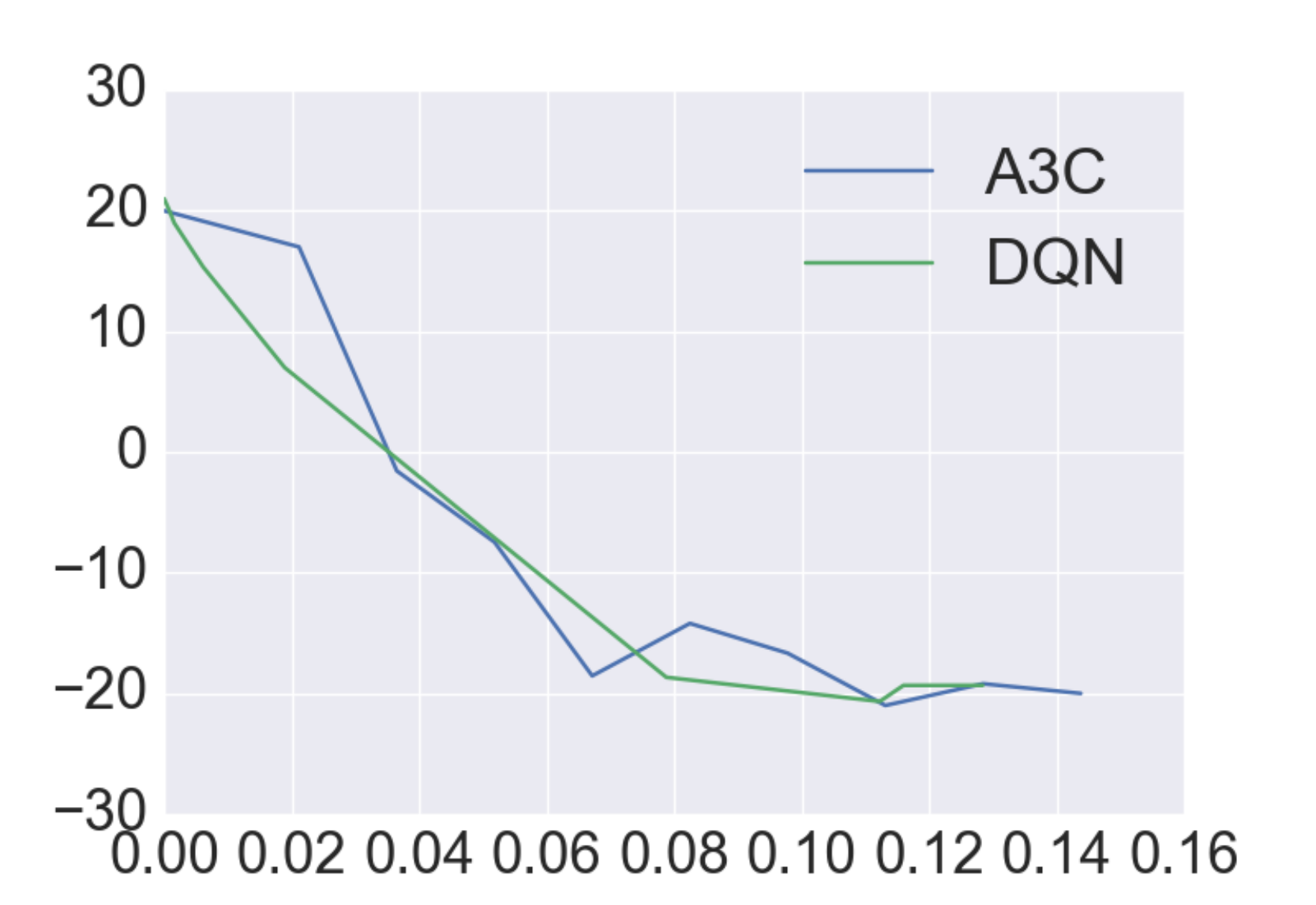}
  \caption{Pong}
  \label{fig:sfig1}
\end{subfigure}%
\begin{subfigure}{.2\textwidth}
  \centering
  \includegraphics[width=1.0\linewidth]{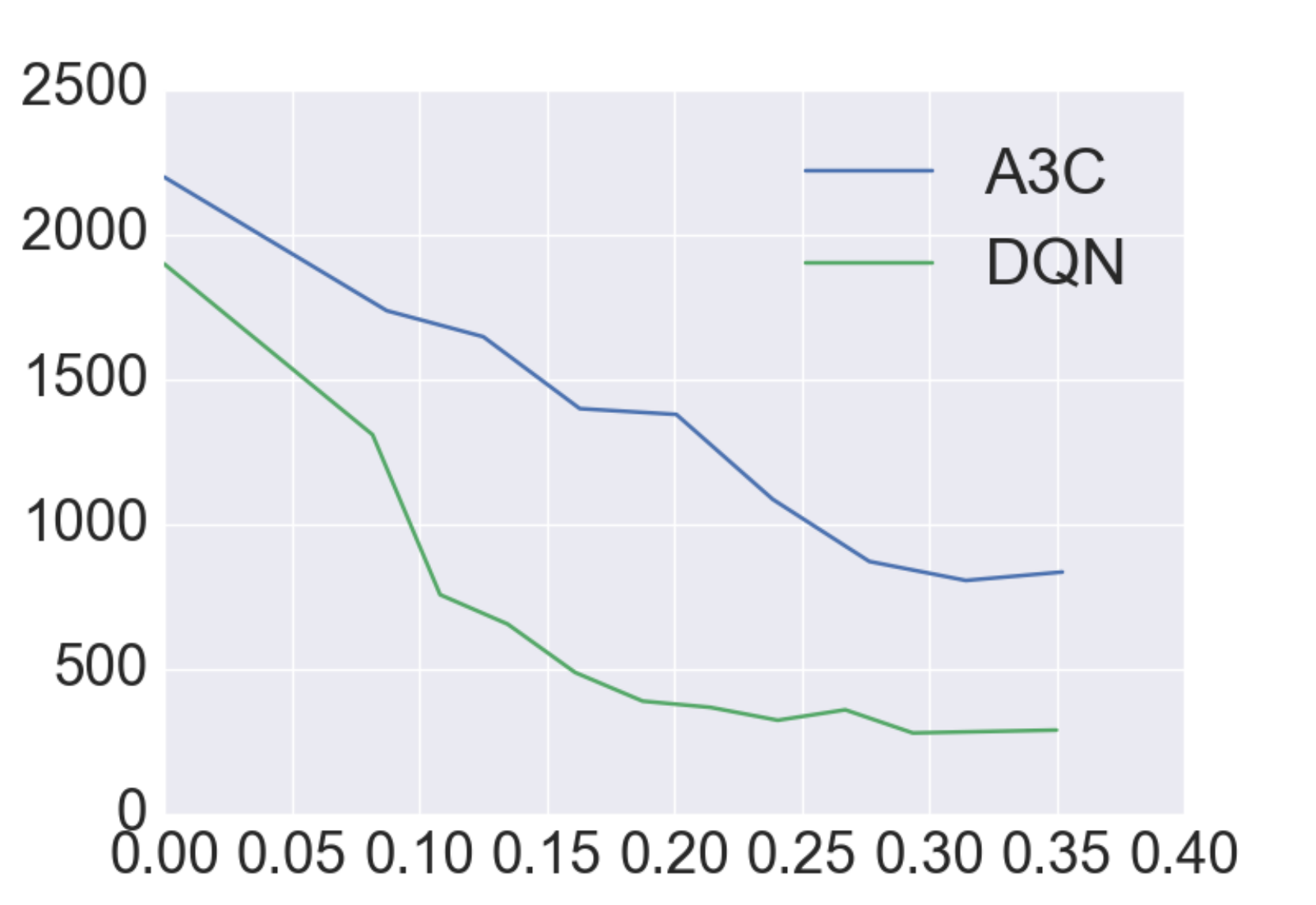}
  \caption{Seaquest}
  \label{fig:sfig2}
\end{subfigure}%
\begin{subfigure}{.2\textwidth}
  \centering
  \includegraphics[width=1.0\linewidth]{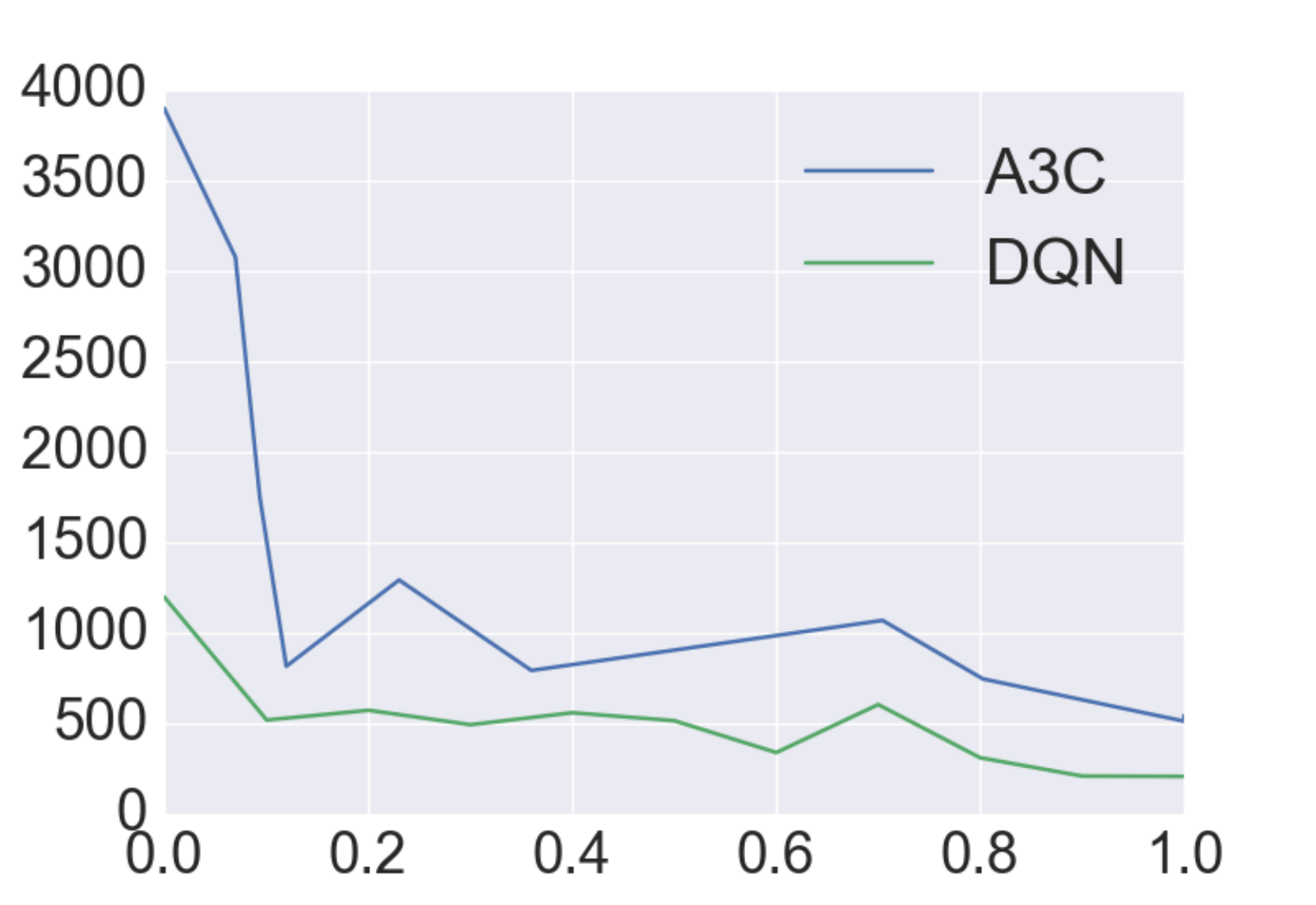}
  \caption{MsPacman}
  \label{fig:sfig3}
\end{subfigure}%
\begin{subfigure}{.2\textwidth}
  \centering
  \includegraphics[width=1.0\linewidth]{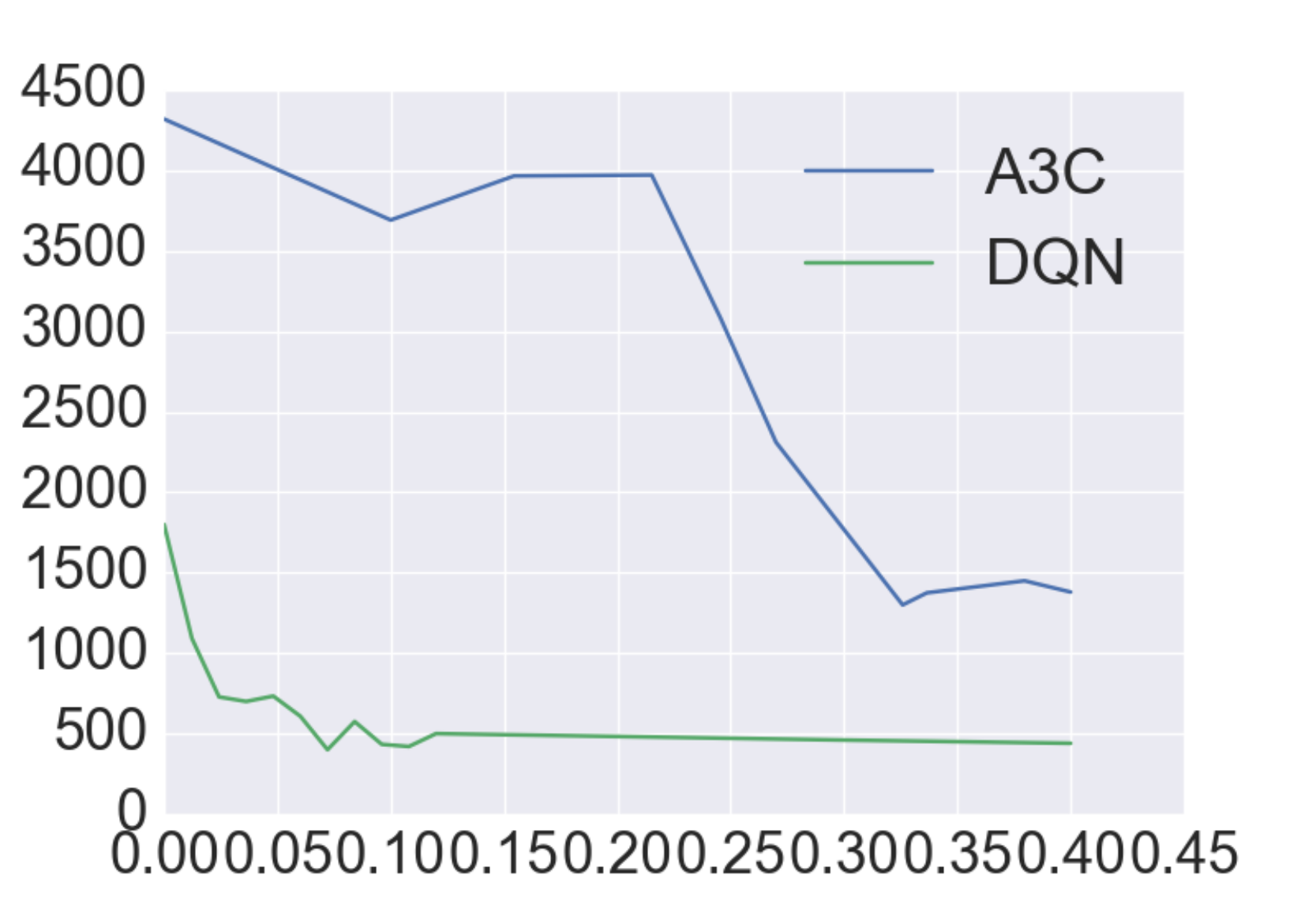}
  \caption{ChopperCommand}
  \label{fig:sfig4}
\end{subfigure}%
\begin{subfigure}{.2\textwidth}
  \centering
  \includegraphics[width=1.0\linewidth]{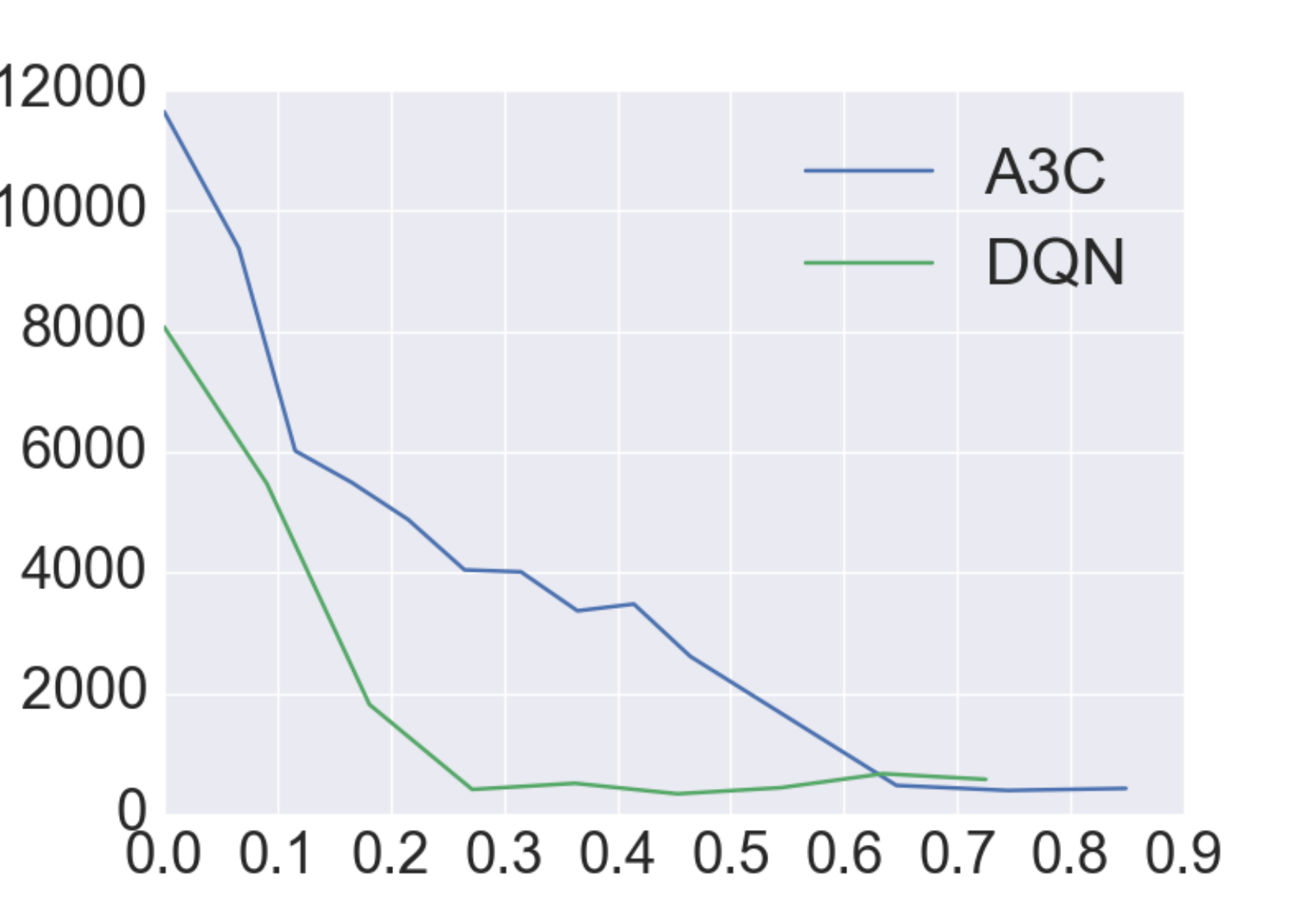}
  \caption{Qbert}
  \label{fig:sfig5}
\end{subfigure}\vspace{-2mm}
\caption{Accumulated reward (y-axis) v.s. Portions of time steps the agent is attacked (x-axis) of Strategically-timed Attack in 5 games. The blue and green curves correspond to results of A3C and DQN, respectively. A larger reward means the deep RL agent is more robust to the strategically-timed attack.}
\label{fig:fig}
\end{figure*}

\begin{figure*}[t!]\vspace{-2mm}
\centering
\begin{subfigure}{.2\textwidth}
  \centering
  \includegraphics[width=1.0\linewidth]{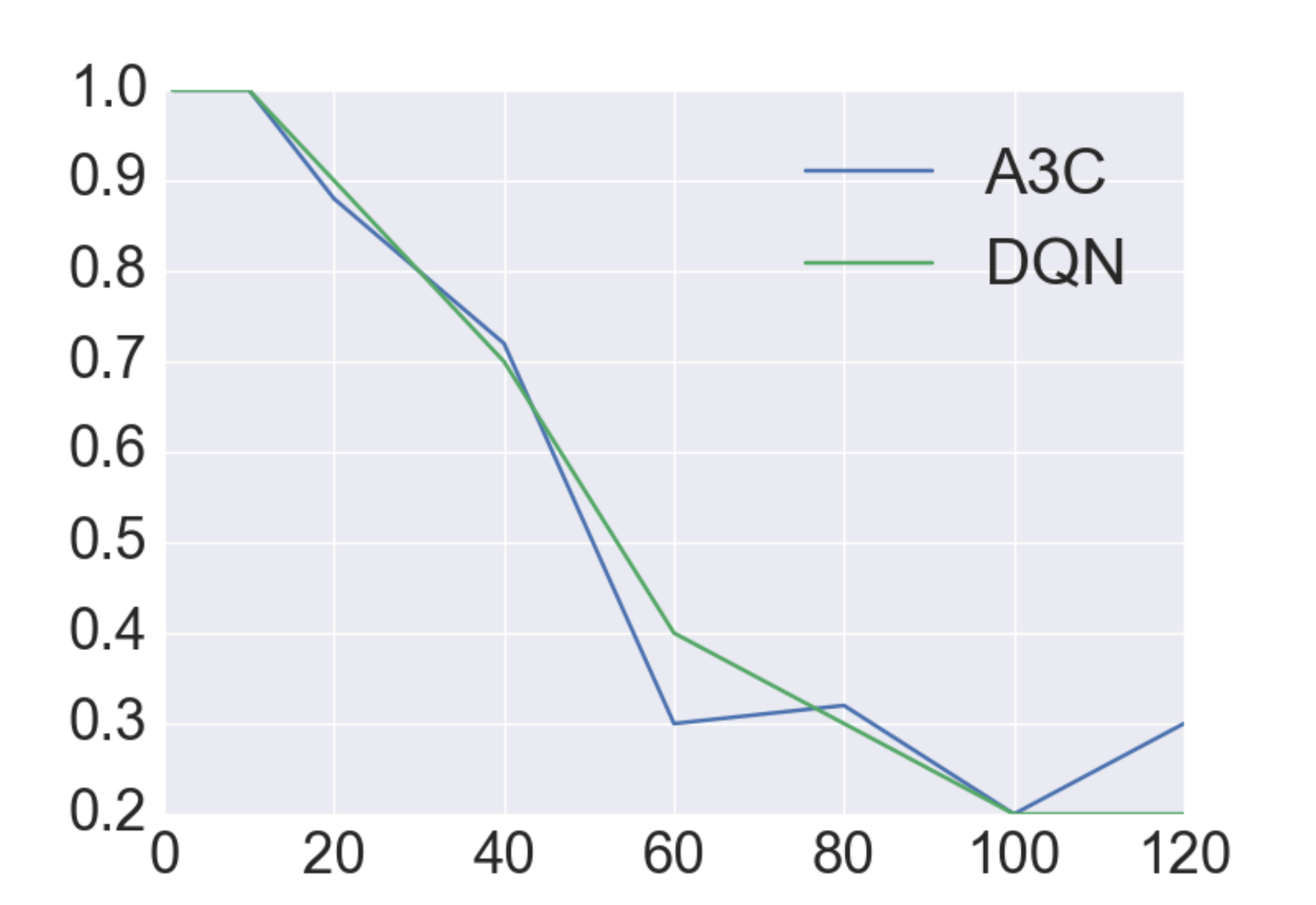}
  \caption{Pong}
  \label{fig:sfig1}
\end{subfigure}%
\begin{subfigure}{.2\textwidth}
  \centering
  \includegraphics[width=1.0\linewidth]{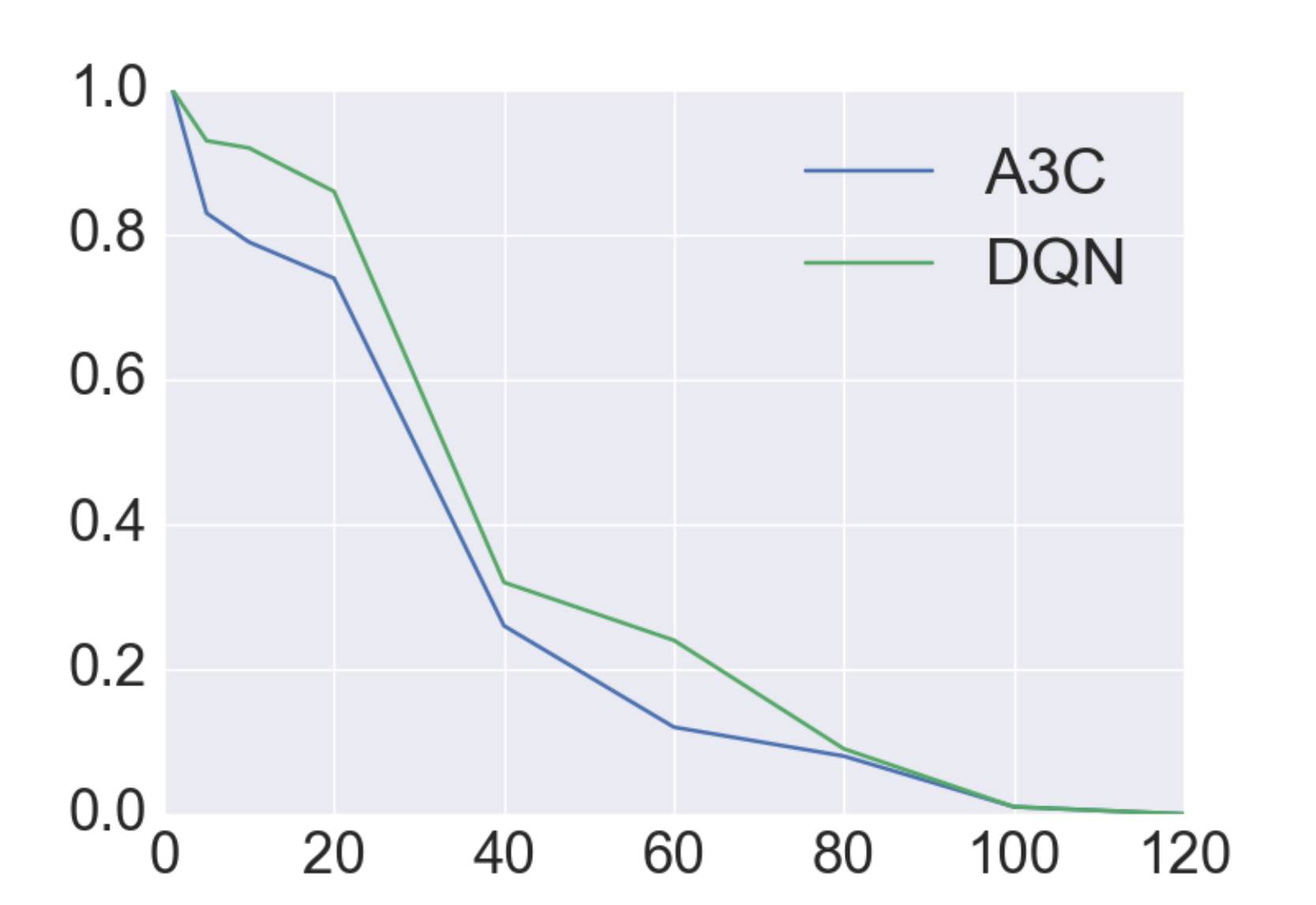}
  \caption{Seaquest}
  \label{fig:sfig2}
\end{subfigure}%
\begin{subfigure}{.2\textwidth}
  \centering
  \includegraphics[width=1.0\linewidth]{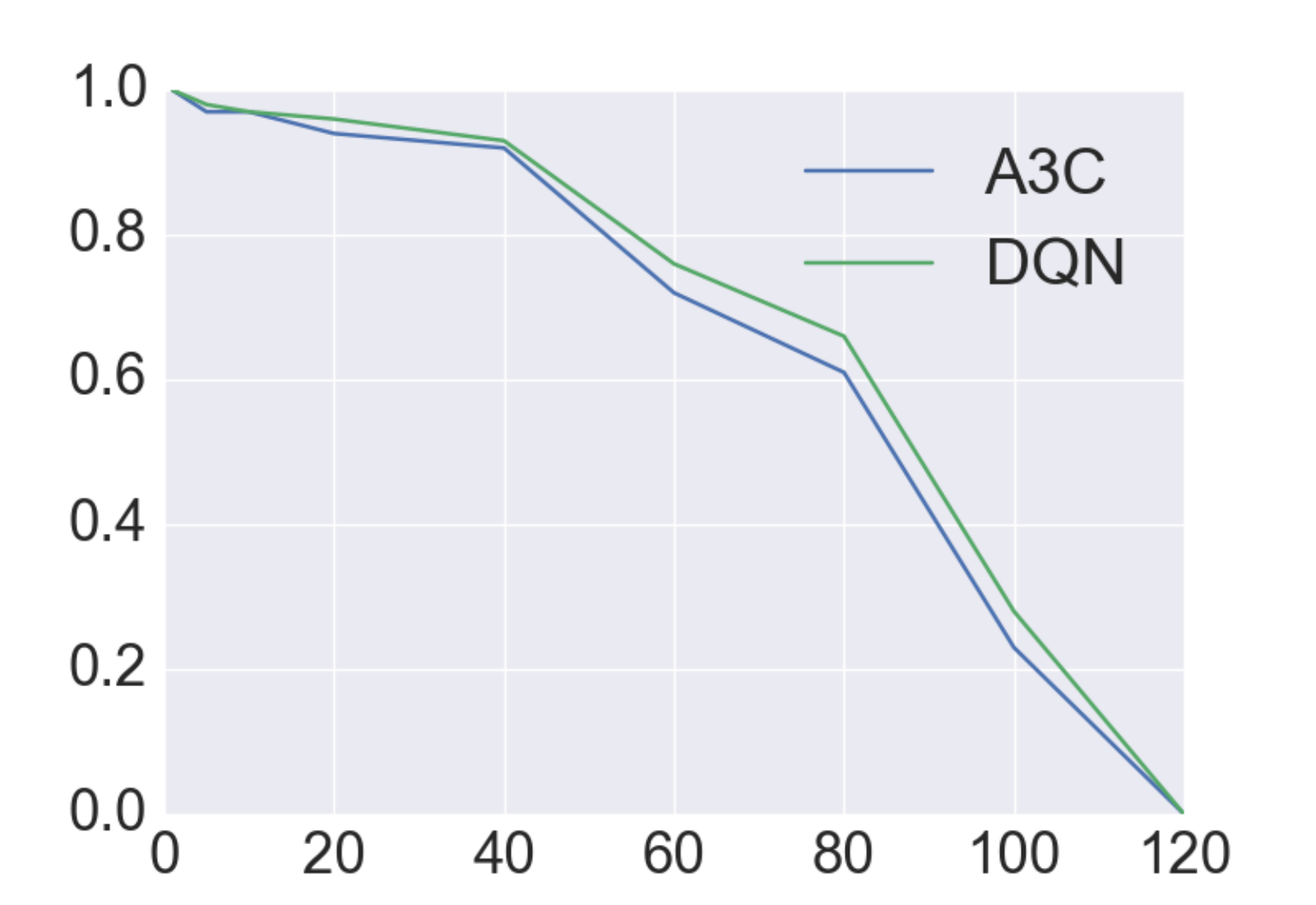}
  \caption{MsPacman}
  \label{fig:sfig3}
\end{subfigure}%
\begin{subfigure}{.2\textwidth}
  \centering
  \includegraphics[width=1.0\linewidth]{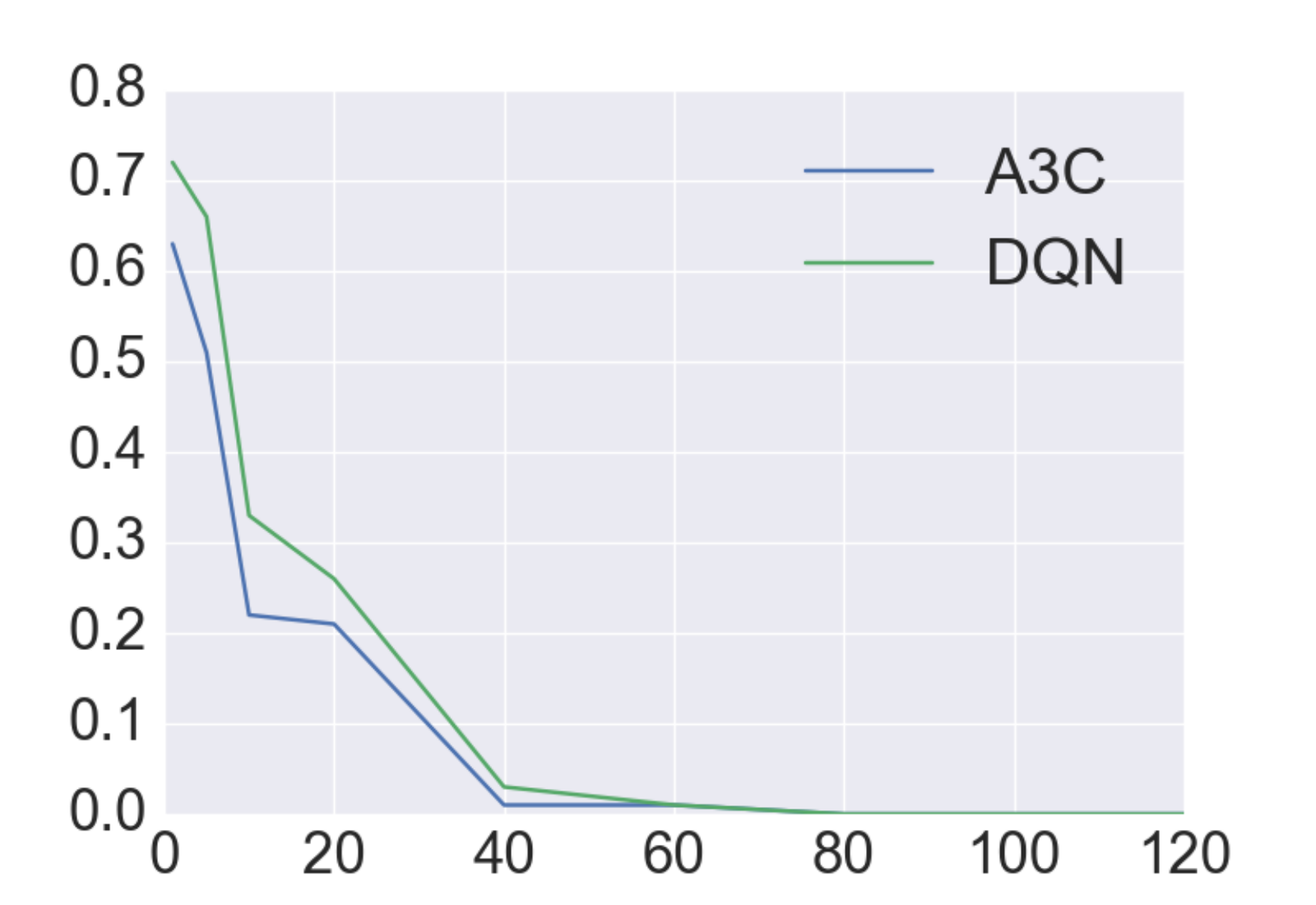}
  \caption{ChopperCommand}
  \label{fig:sfig4}
\end{subfigure}%
\begin{subfigure}{.2\textwidth}
  \centering
  \includegraphics[width=1.0\linewidth]{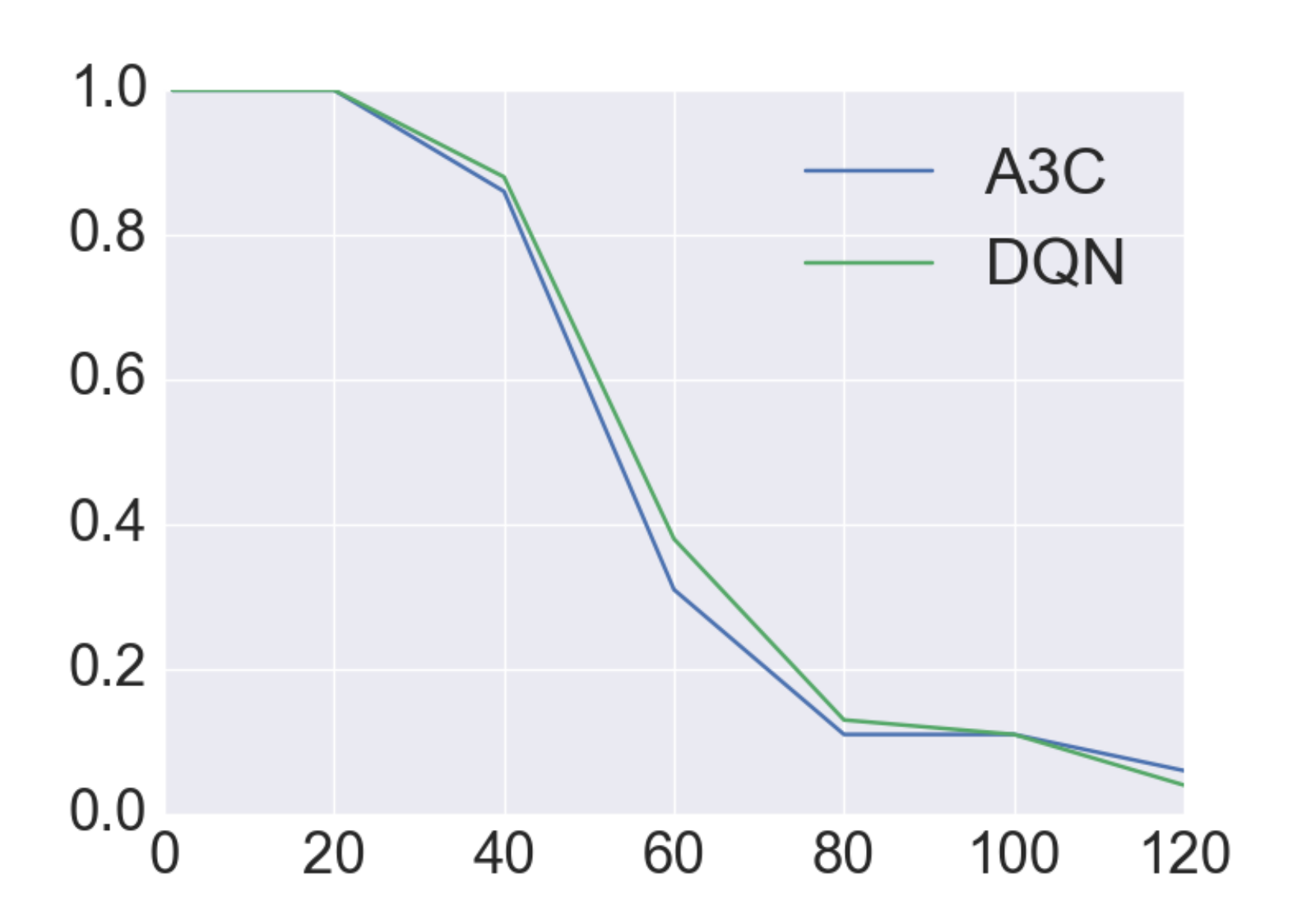}
  \caption{Qbert}
  \label{fig:sfig5}
\end{subfigure}\vspace{-2mm}
\caption{Success rate (y-axis) v.s. $H$ steps in the future (x-axis) for Enchanting Attack in 5 games. The blue and green curves correspond to results of A3C and DQN, respectively. A lower \ignore{success} rate means that the deep RL agent is more robust to the enchanting attack.}
\label{fig:fig2}
\end{figure*}

\subsection{Strategically-Timed Attack}

For each game and for the agents trained by the DQN and A3C algorithms, we launched the strategically-timed attack using different $\beta$ values. Each $\beta$ value rendered a different attack rate, quantifying how often an adversary attacked an RL agent in an episode. We computed the collected rewards by the agents under different attack rates. The results are shown in Fig.~\ref{fig:fig} where the $y$-axis is the accumulated reward and the $x$-axis is the average portion of time steps in an episode that an adversary attacks the agent (i.e., the attack rate).
\sunmincamera{\yenchencamera{}{We show the lowest attack rate until the reward can't get lower. Therefore, the rightmost reward equals the performance of uniform attack (i.e., attack at every time steps).}}{We show the lowest attack rate where the reward reaches the the reward of uniform attack.} From the figure, we found that on average the strategically-timed attack can reach the same effect of the uniform attack by attacking $25\%$ of the time steps in an episode. We also found an agent trained using the DQN algorithm was more vulnerable than an agent trained with the A3C algorithm in most games except Pong. Since the A3C algorithm is known to perform better than the DQN algorithm in the Atari games, this result suggests that a stronger deep RL agent may be more robust to the adversarial attack. This finding, to some extent, echoes the finding in ~\cite{rozsa:towards}, which suggested that a stronger DNN-based recognition system is more robust to the adversarial attack.

\subsection{Enchanting Attack}

The goal of the enchanting attack is to maliciously \yenchenreplace{guide}{lure} the agent toward a target state. In order to avoid the bias of defining target states manually, we synthesized target states randomly. Firstly, we let the agent to apply its policy by $t$ steps to reach an initial state $s_{t}$ and saved this state into a snapshot. Secondly, we randomly sampled a sequence of actions of length $H$ and consider the state reached by the agent after performing these actions as a synthesized target state $s_g$. After recording the target state, we restored the snapshot and run the enchanting attack on the agent and compared the normalized Euclidean distance between the target state $s_{g}$ and the final reached state $s_{t+H}$ where the normalization constant was given by the image resolution.

We considered an attack was successful if the final state had a normalized Euclidean distance to the target state within a tolerance value of 1. To make sure the evaluation was not affected by different stages of the game, we set 10 initial time step $t$ equals to ${[0.0,0.1,...,0.9]} \times L$, where $L$ was the average length of the game episode played by the RL agents 10 times. For each initial time step, we evaluated different $H=\left[1, 5, 10, 20, 40, 60, 80, 100, 120\right]$. Then, for each $H$, we computed the success rate (i.e., number of times the adversary misguided the agent to reach the target state divided by number of trials). We expected that a larger $H$ would correspond a more difficult enchanting attack problem. In Fig.~\ref{fig:fig2}, we show the success rate (y-axis) as a function of $H$ in 5 games. We found that the agents trained by both the A3C and DQN algorithms were enchanted. When $H<40$, the success rate was more than $70\%$ in several games (except Seaquest and ChopperCommand). The reason that enchanting attack was less effective on Seaquest and ChopperCommand was because both of the games include multiple random enemies such that our trained video prediction models were less accurate.




\section{Conclusion}\label{sec.Con}

We introduced two novel tactics of adversarial attack on deep RL agents: the strategically-timed attack and the enchanting attack. In five Atari games, we showed that the accumulated rewards collected by the agents trained using the DQN and A3C algorithms were significantly reduced when they were attacked by the strategically-timed attack even with just $25\%$ of the time steps in an episode. Our enchanting attack combining video prediction and planning can lure deep RL agent toward maliciously defined target states in $40$ steps with more than $70\%$ success rate in 3 out of 5 games. In the future, we plan to develop a more sophisticated strategically-timed attack method. We also plan to improve video prediction accuracy of the generative model for improving the success rate of enchanting attack on more complicated games. 
Another important direction of future work is developing defenses against adversarial attacks. Possible methods including augmenting training data with adversarial examples (as in \cite{goodfellow:explaining}, or training a subnetwork to detect adversarial input at test time and deal with it properly.

\section*{Acknowledgements}\label{sec.Ack}
We would love to thank anonymous reviewers, Chun-Yi Lee, Jan Kautz, Bryan Catanzaro, and William Dally for their useful comments. We also thank MOST 105-2815-C-007-083-E and MediaTek for their support.

\bibliographystyle{named}
\bibliography{ijcai17}

\end{document}